\documentclass[letterpaper]{article} 
\usepackage{aaai24}  
\usepackage{times}  
\usepackage{helvet}  
\usepackage{courier}  
\usepackage[hyphens]{url}  
\usepackage{graphicx} 
\usepackage{natbib}  
\usepackage{caption} 
\usepackage{algorithm}
\usepackage{algorithmic}
\usepackage{tabularx}
\usepackage{epsfig}
\usepackage{graphicx}
\usepackage{amsmath}
\usepackage{amssymb}
\usepackage{euscript} 
\usepackage{multirow}
\usepackage{caption}
\usepackage{color}

\usepackage{newfloat}
\usepackage{listings}
\DeclareCaptionStyle{ruled}{labelfont=normalfont,labelsep=colon,strut=off} 
\floatstyle{ruled}
\newfloat{listing}{tb}{lst}{}
\floatname{listing}{Listing}
\def\abbr{NeRF-VPT}

\title{NeRF-VPT: Learning Novel View Representations with Neural Radiance Fields via View Prompt Tuning}
\author{
    Linsheng Chen\textsuperscript{\rm 1}\equalcontrib,
    Guangrun Wang\textsuperscript{\rm 2}\equalcontrib,\\
    Liuchun Yuan\textsuperscript{\rm 1},
    Keze Wang\textsuperscript{\rm 1}\thanks{Corresponding Author.},
    Ken Deng\textsuperscript{\rm 1}, Philip H.S. Torr\textsuperscript{\rm 2}
}
\affiliations{
    \textsuperscript{\rm 1}Sun Yat-sen University     \\
    \textsuperscript{\rm 2}University of Oxford\\
    \{kezewang, wanggrun,ylc0003\}@gmail.com,\{chenlsh36,dengk26\}@mail2.sysu.edu.cn,philip.torr@eng.ox.ac.uk
}

\usepackage{bibentry}

\begin{document}

\maketitle

\begin{figure*}
    \centering
    \includegraphics[width=\linewidth]{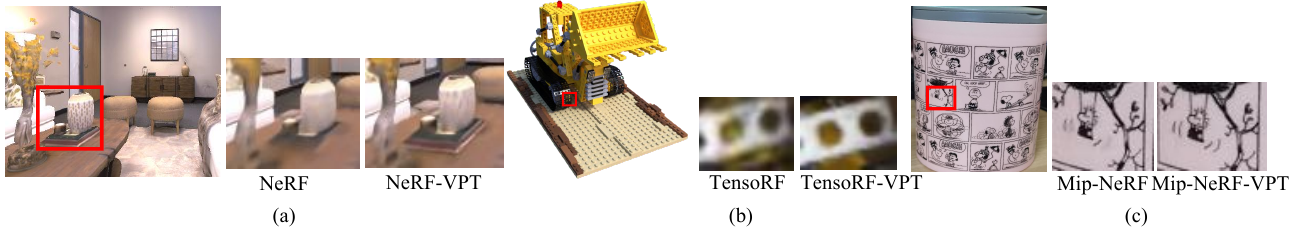}
    \caption{Qualitative comparison between our proposed \abbr\ and its corresponding baselines. (a) is on the Replica dataset. (b) is on the Realistic Synthetic 360 dataset. (c) is on a user-captured scene using a mobile phone.}\label{fig:intro}
\end{figure*}

\begin{abstract}
Neural Radiance Fields (NeRF) have garnered remarkable success in novel view synthesis. Nonetheless, the task of generating high-quality images for novel views persists as a critical challenge.
While the existing efforts have exhibited commendable progress, capturing intricate details, enhancing textures, and achieving superior Peak Signal-to-Noise Ratio (PSNR) metrics warrant further focused attention and advancement.
In this work, we propose \abbr, an innovative method for novel view synthesis to address these challenges.
Our proposed \abbr\ employs a cascading view prompt tuning paradigm, wherein RGB information gained from preceding rendering outcomes serves as instructive visual prompts for subsequent rendering stages, with the aspiration that the prior knowledge embedded in the prompts can facilitate the gradual enhancement of rendered image quality.
\abbr\ only requires sampling RGB data from previous stage renderings as priors at each training stage, without relying on extra guidance or complex techniques. Thus, our \abbr\ is plug-and-play and can be readily integrated into existing methods.
By conducting comparative analyses of our \abbr\ against several NeRF-based approaches on demanding real-scene benchmarks, such as Realistic Synthetic 360, Real Forward-Facing, Replica dataset, and a user-captured dataset, we substantiate that our \abbr\ significantly elevates baseline performance and proficiently generates more high-quality novel view images than all the compared state-of-the-art methods. Furthermore, the cascading learning of \abbr\ introduces adaptability to scenarios with sparse inputs, resulting in a significant enhancement of accuracy for sparse-view novel view synthesis. The source code and dataset are available at \url{https://github.com/Freedomcls/NeRF-VPT}.
\end{abstract}

\section{Introduction}

In recent years, Neural Radiance Fields (NeRF) \cite{mildenhall2021nerf} have exhibited significant promise across a range of disciplines, including novel view synthesis \cite{tretschk2021non,tao2023lidar}, 3D reconstruction \cite{barron2021mip,verbin2022ref,mildenhall2022nerf}, and 3D editing \cite{kania2022conerf}. NeRF employs a neural network to encapsulate the color and density attributes inherent to each point within 3D space. This encoding technique enables the synthesis of novel views of a scene, encompassing a comprehensive range of viewpoints. Nonetheless, as visually exemplified in Figure \ref{fig:intro}., generating high-quality images portraying novel views using NeRF remains challenging.

In recent times, a plethora of outstanding work endeavors have emerged to address this challenge, yielding encouraging results \cite{barron2021mip,roessle2022dense,varma2022attention, bautista2022gaudi,vora2021nesf,johari2022geonerf,tschernezki2022neural,kurz2022adanerf, arandjelovic2021nerf,hedman2021baking,VolSDF,tancik2022block}. While the strides made by current endeavors are considerable, there persists a clear necessity for a more intensified concentration on facets including the precise capture of intricate details \cite{bautista2022gaudi,vora2021nesf,kurz2022adanerf}, refinement of textures \cite{roessle2022dense,vora2021nesf,kurz2022adanerf}, and attainment of elevated Peak Signal-to-Noise Ratio (PSNR) metrics \cite{barron2021mip,roessle2022dense,varma2022attention,johari2022geonerf,kurz2022adanerf}.

In this paper, we introduce \abbr, a method that incorporates effective guidance to fully harness the inherent potential of NeRF-based models, resulting in more comprehensive and accurate outcomes.
We adopt a cascading view prompt tuning paradigm, where RGB information obtained from preceding rendering iterations acts as informative visual prompts for subsequent rendering stages. The underlying goal is to leverage the prior knowledge embedded in these prompts to progressively enhance the quality of the rendered images.
Our method substantially augments the guidance for both the reconstruction and rendering processes within the neural radiance field framework. It excels at iteratively combining comprehensive RGB information and spatial dependencies through a multi-stage refinement process, capitalizing on the benefits of cascading learning strategies \cite{lin2017recurrent, ramakrishna2014pose}.

As illustrated in Figure ~\ref{fig:intro}., our proposed NeRF-VPT has been meticulously designed to exhibit modularity and portability, thereby facilitating effortless integration with prior NeRF techniques such as vanilla NeRF, Mip-NeRF \cite{barron2021mip}, and TensoRF \cite{chen2022tensorf}. This integration serves to further augment the performance of the prior NeRF approaches.
The main contributions of our work are summarized as follows:

\begin{itemize}
    \item We introduce a novel and efficacious framework, referred to as \abbr, featuring a cascading view prompt tuning designed to iteratively leverage the representation inherent to the NeRF family in the form of visual prompts.
    \item \abbr\ stands out for its inherent simplicity and seamless integration capabilities with pre-existing models, thereby elevating the performance of previous models to achieve heightened levels of excellence. Our method can be plug-and-play, which can be integrated into prior NeRF methods, such as vanilla NeRF, Mip-NeRF, and TensoRF.
    \item Through qualitative and quantitative experiments, we demonstrate that our method effectively improves the performance of view synthesis.
\end{itemize}

In summary, our approach offers a promising solution for enhancing the performance of NeRF-based view synthesis. Furthermore, our framework is plug-and-play, capable of augmenting the capabilities of existing NeRF methods.

\section{Related Work}

The NeRF architecture learns 3D scene representation through supervised learning from multi-view RGB images. It employs direct rays for pixel rendering and equidistant space sampling. NeRF's success in image synthesis has spurred significant research interest, resulting in various methods aiming to enhance its visual quality.

\subsubsection{Priors in Scene Representation.} RGB images alone lack comprehensive structural scene information, necessitating the incorporation of additional priors to guide the learning process. Recent advancements emphasize the efficacy of depth information as a valuable prior for refining scene representation \cite{deng2022depth,roessle2022dense}. Depth priors offer a suitable optimization direction, particularly during initial training stages, resulting in significant performance enhancements. Other studies \cite{chen2022structnerf, varma2022attention, bautista2022gaudi} have highlighted NeRF's limitations in adhering to specific spatial constraints, such as planar or the epipolar geometry. To mitigate these shortcomings, these approaches have embedded relative information within the NeRF framework, leading to improved consistency. These methods involve additional loss functions and constraints during optimization. However, these approaches utilizing prior knowledge have limitations in their broader applicability. They often rely on external networks or necessitate specific data preprocessing steps. In contrast, our proposed method capitalizes exclusively on the network's internal output for prior knowledge, enhancing its potential for transferability and wider utility.

\subsubsection{Improvements on Rendering.} The rendering process plays a pivotal role in obtaining RGB images for novel views. Previous research endeavors \cite{zhi2021place, tian2021semantic, tschernezki2022neural, guo2022neural} have predominantly concentrated on enhancing integration functions and devising more effective sampling methods. Acknowledging the challenge posed by single straight rays in capturing the influence of neighboring points, \cite{barron2021mip} introduced Mip-NeRF. This innovative approach substitutes a single ray with 3D conical frustums, yielding anti-aliased and visually crisp images in view synthesis. Furthermore, NeRF encounters limitations in terms of rendering's sampling effectiveness, which can restrict the optimization's coverage. To tackle these issues, approaches like those presented by \cite{kurz2022adanerf} and Arandjelović and Rieger \cite{arandjelovic2021nerf} incorporate predictive networks to gauge the sampling's significance. This strategy reduces computational complexity in rendering, allowing for more resources to be allocated to expanding networks and enhancing result quality.
Although these efforts have made significant strides, there remains a need for greater focus on capturing intricate details, refining textures, and achieving higher PSNR metrics.

\begin{figure*}[t]
    \centering
    \includegraphics[width=0.75\linewidth]{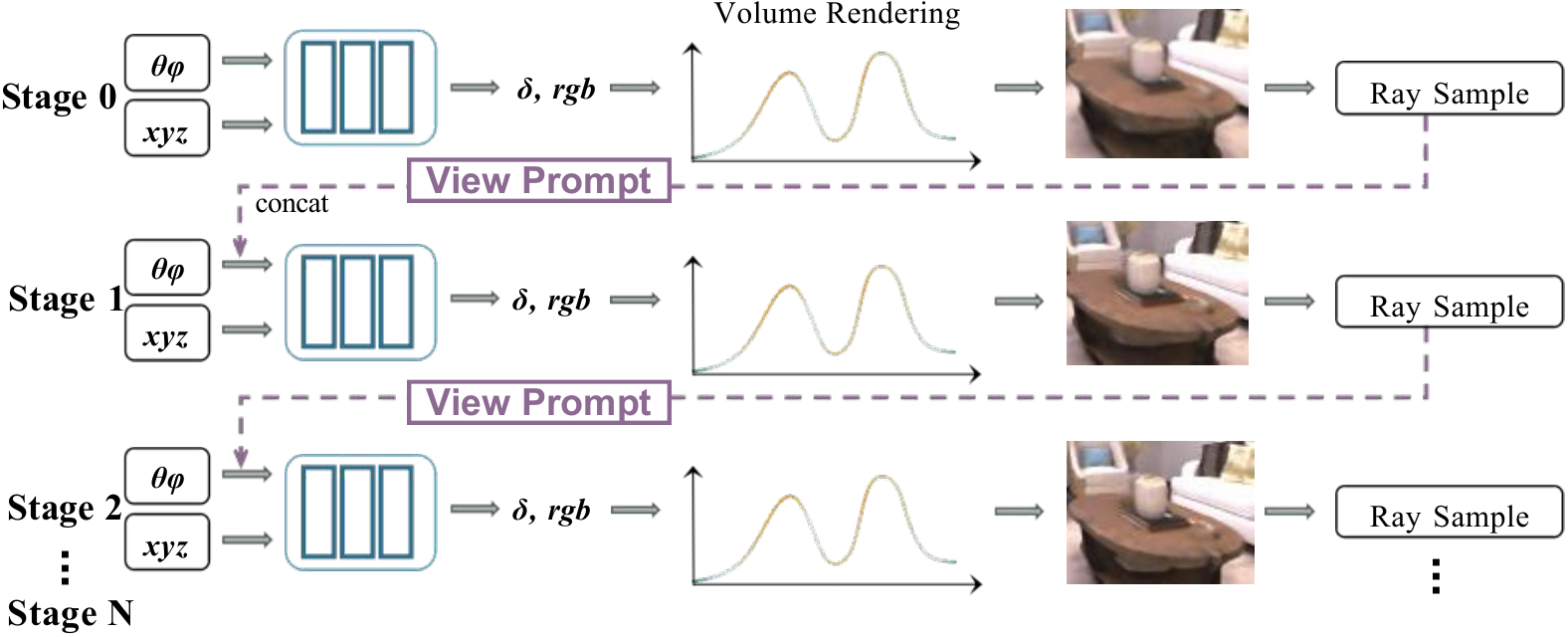}
    \caption{The overview of our proposed framework. In each iteration, the image rendered in the previous round is combined with the encoded direction after ray sampling, where stage 0 is a NeRF that does not rely on additional input. Each stage adopts a new network to learn with the previous RGB information.}\label{fig:supernerf}
\end{figure*}

\subsubsection{Advanced Structure.} Excellent endeavors strive to elevate the photorealistic synthesis of novel views through the incorporation of sophisticated architectures, such as Recursive-NeRF \cite{yang2022recursive} and BungeeNeRF \cite{xiangli2022bungeenerf}. These approaches aim to partition the 3D space into more intricate or finely-grained segments, each corresponding to an individual network. This partitioning allows the scene to be represented across multiple levels of granularity, rendering it well-suited for encompassing diverse and intricate scenarios, such as natural landscapes, urban streets, and indoor environments.

Our approach sets itself apart from Recursive-NeRF and BungeeNeRF through the utilization of an innovative defective visual prompt tuning technique. This technique enables the integration of prior knowledge derived from earlier rendered outcomes, thereby serving as valuable guidance for the network to gain a deeper understanding of intricate scenes. This, in turn, contributes to an enhancement in the quality of novel view synthesis.

\subsubsection{Visual Prompt Tuning.}
Visual prompt learning \cite{jia2022visual} capitalizes on an extensive corpus of scholarly investigations encompassing domains such as visual question answering \cite{chappuis2022prompt}, few-shot learning \cite{alayrac2022flamingo}, and generative models \cite{wu2022generative}. Through the amalgamation of insights harvested from these scholarly realms, the field of visual prompt learning aspires to engender innovative methodologies for the generation of responses to visual inputs, empowering them to grasp and engender content from visual stimuli more adeptly.
Our approach to prompt tuning significantly diverges from prevailing methods in the realm of visual prompt tuning. Specifically, we introduce a novel cascading view prompt tuning paradigm that embeds prior knowledge within our visual prompts. In this paradigm, RGB information derived from antecedent rendering iterations assumes the role of informative visual prompts for subsequent rendering stages.

\section{Methodology}

In this section, we commence by laying the groundwork with foundational concepts. We then proceed to the structure of \abbr. Following this, we elucidate the intricacies of the \abbr\ design, underscoring its effectiveness as a versatile and comprehensive framework.

\subsection{Preliminary}

For a point in scene $\mathbf{x} \in \mathbb{R}^3$ and viewing direction $\mathbf{d} \in \mathbb{S}^2$, NeRF \cite{mildenhall2021nerf} learns a function to query each point color $\mathbf{c}$  and voxel density ${\sigma}$ with specific views: 
\begin{small}\begin{equation}
(\mathbf{c}, \sigma) = \mathit{F}(\mathbf{x}, \mathbf{d}).
\label{eq:nerf_f}
\end{equation}\end{small}Specifically, the $\mathbf{c}$ and $\sigma$ are output after $\mathit{F_c}$ and $\mathit{F_d}$. 
In NeRF, $\mathbf{x}$  and $\mathbf{d} $ will present as position encoding and direction encoding:\begin{small}\begin{equation}
\begin{aligned}
\mathit{\gamma(v)} = & (sin(2^0\pi v), cos(2^0\pi v), sin(2^1\pi v), cos(2^1\pi v), \\
& ...,  sin(2^{N-1}\pi v), cos(2^{N-1}\pi v))
\end{aligned}.
\label{eq:nerf_enc}
\end{equation}\end{small}Volumetric rendering is proposed to create a 2D image based on $\mathit{F}$. It imagines each pixel emitting a single ray to 3D space from the image center $\mathbf{o}$, and sampling points along this ray, $\mathbf{x_i} =  \mathbf{o} + {t_i}\mathbf{d}$. Thus, the color of this pixel can be approximated by numerical quadrature:\begin{small} 
\begin{equation}
\mathbf{\hat{c}(r)} = \int_{t_n}^{t_f} \mathit{T}(t)\sigma(t)\mathbf{c}(t) dt;  \mathit{T}(t) = e^{-\int_{t_n}^{t}\sigma(s)ds}, 
\label{eq:nerf_render}
\end{equation}\end{small}where $\mathit{T}$ handles the occlusion; ${t_n}$, ${t_f}$ denote the near and far boundary of the 3d scene. Limited to computing resources, it adopts a coarse-to-fine sampling. 
The rendered color of each ray corresponds to the ground-truth value, thus the training is supervised by:
\begin{equation}
\mathcal{L} = \sum_{{r} \in \mathcal{R}} \vert\vert \mathbf{\hat{c}(r)} - \mathbf{c(r)} \vert\vert, 
\label{eq:nerf_loss}
\end{equation}where $\mathcal{R}$ present all image pixels in training data, and $\mathbf{c(r)}$ is ground-truth color.

Priors within the scene can optimize sampling efficiency, offer an enhanced methodology for characterizing occlusion, and introduce supplementary loss functions in specific conditions.
These priors are typically incorporated alongside the position $\mathbf{x}$. Consequently, the mapping function becomes: $(\mathbf{c}, \sigma) = \mathit{F}(\mathbf{x}, {d}, {p})$, where ${p}$ represents priors. The advancements in rendering primarily involve updates to Eq. \eqref{eq:nerf_render}, aimed at achieving more realistic results. These improvements can be implemented in various ways, rendering it intricate to consolidate them into a single comprehensive equation. Furthermore, these NeRF variants with hierarchical structures adopt a repetitive Eq. \eqref{eq:nerf_f}: $\lbrace (\mathbf{c_i}, \sigma_i) = \mathit{F_i}(\mathbf{x_i}, {d}) | i\in N \rbrace$. Some methods not only focus on a single improvement but also apply them synchronously.

\subsection{Structural Design of \abbr}

Most NeRF-based methods optimize a scene from random initialization. However, we believe that the inclusion of prior knowledge encompassing specific scene information could potentially simplify the model's comprehension of the scene.
Hence, our objective is to enable the network to directly learn the posterior distribution related to prior knowledge. 
To this end, we propose \abbr, a method that leverages prior knowledge sampled from the RGB space as a view prompt to aid the model in understanding the scene. Moreover, we introduce cascading learning to progressively enhance the quality of synthesized images. Figure ~\ref{fig:supernerf}. illustrates the framework of our approach.

\subsection{View Prompt with Prior Knowledge}  
To incorporate scene information as prior knowledge embedded by a view prompt into the network, 
we conduct the following experiments: Initially, we attempt to extract features from multiple images in the scene and performed feature fusion by Swin-Transformer or another CNN-based network. The fused features are then incorporated into an MLP (Multi-Layer Perceptron) using the SIREN activation function. However, the experimental results show limited improvement, primarily because it is challenging to encapsulate multi-view information within a single feature. Furthermore, we explore the direct provision of ground truth images from the training set to the network during the training process for each viewpoint. This approach can be seen as providing the network with an answer during the learning process. The experimental results demonstrate a significant improvement in the quality of synthesized images during both the training and validation stages. However, this approach also introduces a challenge: we are unable to provide ground truth images for unknown viewpoints during the testing phase as prior knowledge.

In order to address this issue, we first employ NeRF to reconstruct the scene and render images from training, validation, and testing viewpoints. We incorporate this information as prior knowledge in the form as a prompt for the neural radiance field model for retraining, ensuring that the views during training, validation, and testing remain consistent with the NeRF training process. This training phase is fundamentally a guiding procedure facilitated through view prompt tuning (see Figure \ref{fig:supernerf}). Compared to directly using ground truth as priors, it can be understood as providing answers that may not be particularly accurate but effectively guide the optimization direction during the learning process. This approach can somewhat reduce the difficulty of learning while still benefiting from the supervision of ground truth. Importantly, during the testing phase, we can also sample images with an equivalent level of detail as prior knowledge in the form of a prompt.

As shown in Eq. \eqref{eq:nerf_f} and \eqref{eq:nerf_render},  we obtain three essential components for a new view: RGB values, the color, and density of all sampling points in this scene. Among these, RGB values are the most informative, as they encode both the density and color of each ray. Moreover, incorporating RGB values as a prompt brings negligible complexity due to its lower dimensions. Experiments have shown that incorporating pixel values in this way leads to improved performance in novel view synthesis compared to using just the scene color and density information. 

\subsection{View Prompt Tuning with Cascading Learning}

Cascade structures have been proven useful in various computer vision applications, such as object detection \cite{cai2018cascade}, super-resolution \cite{saharia2022image}, and pose estimation \cite{lin2017recurrent}. However, few works in the context of NeRF and beyond have made use of their advantages. Previous methods have focused on upgrading each part of NeRF, while \abbr\ disentangles the generation problem by refining it through a series of stages, sharing the merit of prompt tuning. One of the cornerstones of our method is to re-utilize the representation of the framework.
For clarity, we employ a combination of conventional NeRF and view prompt tuning for illustrative purposes. It is important to highlight that our methods are plug-and-play for seamless integration with other NeRF variants, as demonstrated in our experiments.

 We employ multi-steps to learn the representation of 3D scenes. In the first step, we use NeRF to gain knowledge of 3D space, which can generate the prior (value of RGB) as a view prompt for the next stage. As this step does not include any prior, \abbr\ also calls it the zero stage (stage 0 shown in Figure \ref{fig:supernerf}.). 
 In the second and next steps, we insert the RGB as a view prompt and concatenate it to the direction encoding.
 The recurrent process can be formulated as:\begin{small}\begin{equation}
\begin{aligned}
(\mathbf{c_0}, \sigma_0) &= \mathit{F_0}(PE(\mathbf{x}), DE(\mathbf{d})); \\
(\mathbf{c_i}, \sigma_i) &= \mathit{F_i}(PE(\mathbf{x}), (DE(\mathbf{d}), \mathbf{c(r)_{i-1}}))
\end{aligned}
\label{eq:our_f}
\end{equation}\end{small}
where $\mathit{F_0}$ and  $\mathit{F_i}$ note the mapping function in iterations. 

We explore different methods to connect each stage, incorporating it into position encoding or directly taking it as raw input. We suppose that the encoding is signal decomposition, enabling NeRF to learn high-frequency information. The observation will significantly affect images during synthesis. Therefore, \abbr\ concatenates pixel value after direction encoding as the color is exactly high-frequency data. Moreover, the ablation studies prove the importance of connection with previous stages.

\section{Experiments}
The experiments are divided into the following parts: i) quantitatively and qualitatively showing that our method outperforms prior work; ii) verifying the plug-and-play capability; iii) verifying the adaptability to Sparse-View Novel View Synthesis \cite{wang2023sparsenerf,wang2023perf}; iv) conducting a comprehensive ablation study, including depth comparison, the effect of prior knowledge and effect of prior model re-utilize.

\subsection{Datasets}

\textbf{Realistic Synthetic 360 Dataset.} This dataset consists of 4 Lambertian objects with simple geometry. Each object is rendered at 512x512 pixels from viewpoints sampled on the upper hemisphere. This dataset also consists of 8 objects of complicated geometry and realistic non-Lambertian materials. The resolution used for training is 400 $\times$ 400.

\textbf{Real Forward-Facing Dataset \cite{mildenhall2019local}.} This dataset consists of 8 forward-facing scenes captured with a cellphone at a size of 1008x756 pixels. Following the commonly used protocols \cite{mildenhall2021nerf}, we adopt 504 × 376 resolution for this dataset.

\textbf{User-captured Dataset.} Compiled by capturing footage using a mobile phone, encompassing scenes with complex textures. We adopt 504 × 376 resolution for training.

\textbf{Replicate Dataset \cite{JulianStraub2019TheRD}.} This dataset has high-quality reconstructions of various indoor spaces. Each reconstruction has clean dense geometry, high resolution, and high dynamic range textures, glass, and mirror surface information. The resolution used for training is 320x240.

\textbf{Metrics.} We report the mean of PSNR, structural similarity index (SSIM) \cite{wang2004image}, and the LPIPS \cite{zhang2018unreasonable} perceptual metric for all evaluations.

\textbf{Experimental Settings.}  During the training process, we begin by rendering images from a pre-trained NeRF model (or other NeRF-based models) for the training, validation, and testing viewpoints, which serve as prior knowledge for \abbr. In each subsequent training stage, the model relies solely on its output from the previous iteration as guidance. Loading weights from the previous training round is an option to facilitate faster convergence. Our experiments show that five to six stages on a standard baseline can achieve maximum performance, and training is concluded once this limit is reached. Unless otherwise specified, the following performance comparisons of \abbr\ are based on the model iterated to its performance upper limit.

\begin{table}
    \centering
    \bgroup
    \small
    \resizebox{1.0\linewidth}{!}{
    \setlength{\tabcolsep}{1pt}
    \begin{tabular}{l@{\hskip 1.5pt}|@{\hskip 1.5pt}c@{\hskip 1.5pt}c@{\hskip 1.5pt}c@{\hskip 1.5pt}|@{\hskip 1.5pt}c@{\hskip 1.5pt}c@{\hskip 1.5pt}c@{\hskip 1.5pt}@{\hskip 1.5pt}c@{\hskip 1.5pt}}
    \hline
     & \multicolumn{3}{c@{\hskip 1.5pt}|@{\hskip 1.5pt}}{Realistic Synthetic 360} & \multicolumn{3}{c@{\hskip 1.5pt}@{\hskip 1.5pt}}{Real Forward-Facing} \\
     & PSNR$\uparrow$ & SSIM$\uparrow$ & LPIPS$\downarrow$ & PSNR$\uparrow$ & SSIM$\uparrow$ & LPIPS$\downarrow$ \\
    \hline
    \hline
    NeRF     & 31.01 & 0.947 & 0.081 &  26.50 & 0.811 & 0.250 \\
    IBRnet      &  28.14 & 0.942 & 0.072  & 26.73 & 0.851 & 0.175   \\
    Recursive-NeRF                     &  31.34 & 0.953 & 0.052 & - & -& - \\
    Mip-NeRF     &  33.09 & 0.961 & 0.043  &  - & - & -  \\
    TensoRF          & 33.14   & 0.963 & 0.044  & 27.13  &0.839 &0.164\\
    \hline
    Ours Stage 1           &  34.22 & 0.972  & 0.047   &  27.89 & 0.869  & 0.161   \\
    Ours Stage 6          &  \textbf{34.95}     & \textbf{0.977}     & \
    \textbf{0.042}      &  \textbf{28.95}     & \textbf{0.872}     & \textbf{0.158}     \\
    \hline
    \end{tabular}
    }
    \egroup
    \caption{
    Quantitative comparison to baselines. A quantitative comparison of NeRF-VPT and baseline algorithms on the Realistic Synthetic 360 dataset and Real Forward-Facing dataset. `-' denotes the results are not available. Methods: NeRF~\cite{mildenhall2021nerf}, IBRnet~\cite{wang2021ibrnet}, Recursive-NeRF~\cite{Yang_Zhou_Peng_Liang_Mu_Hu_2022}, Mip-NeRF~\cite{barron2021mip}, TensoRF~\cite{chen2022tensorf}.
    }\label{tab:results}
\end{table}

\subsection{Comparison with State-of-the-Art Methods}
\begin{figure*}
    \centering
    \includegraphics[width=1\linewidth]{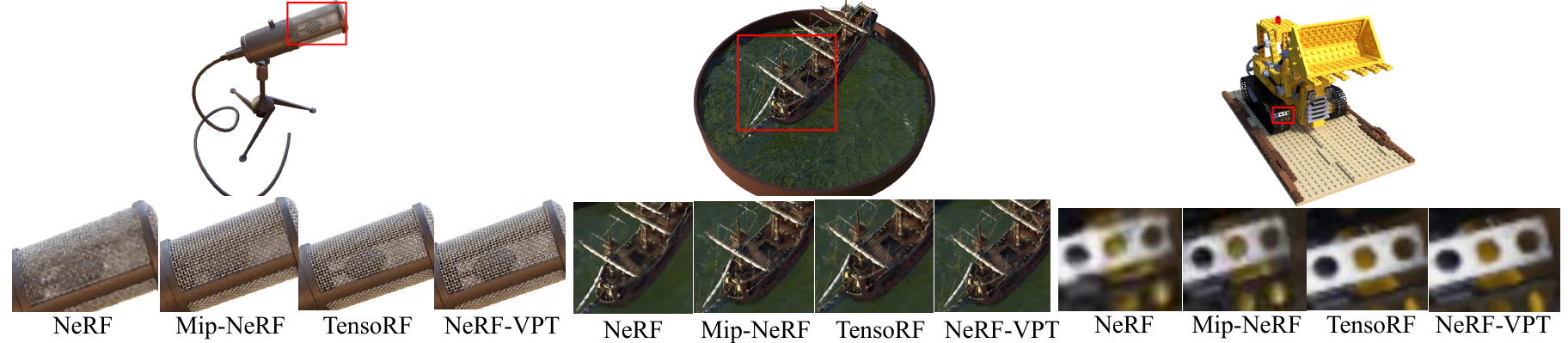}
    \caption{Qualitative results of \abbr\ and comparison methods (NeRF, Mip-NeRF, TensoRF) on Realistic Synthetic 360 Dataset. Please zoom in for better viewing.}\label{fig:compare}
\end{figure*}

\begin{figure}
    \centering
    \footnotesize
    \includegraphics[width=1.0\linewidth]{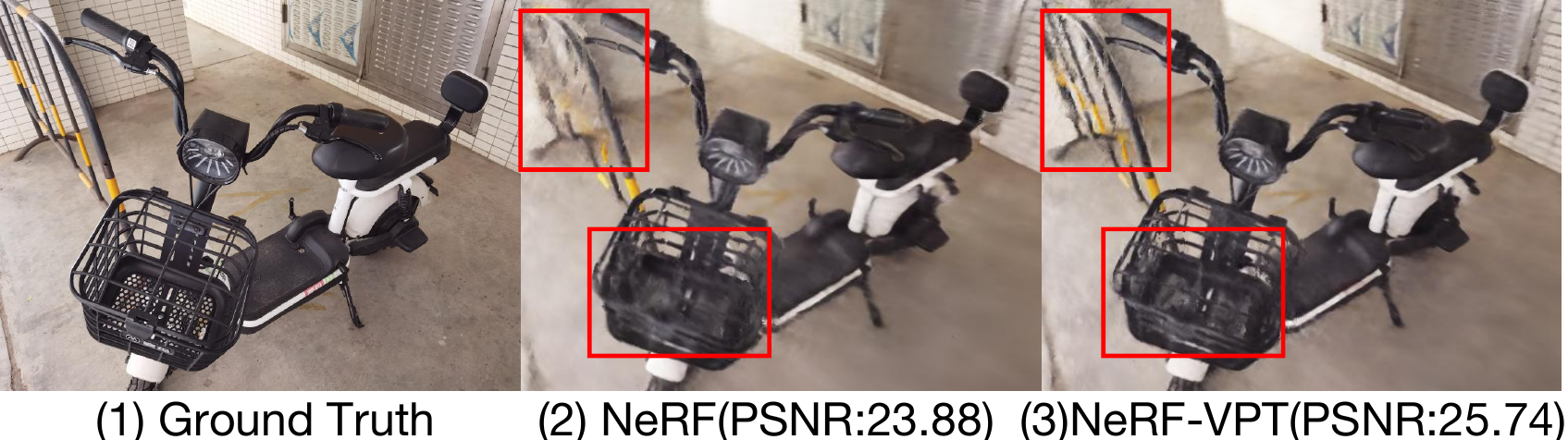}
    \caption{Visual comparisons on real-world datasets.}\label{fig:motor}
\end{figure}

We compare our approach to several current state-of-the-art methods: NeRF, IBRnet, Recursive-NeRF, Mip-NeRF, and TensoRF. 
Neural Radiance Fields (NeRF) is a well-established method that employs Multi-Layer Perceptrons (MLPs) to implicitly represent scenes by utilizing encoded position and direction as input. IBRnet enhances the generalization capability of multi-scene rendering by learning from multi-view images. Mip-NeRF is an enhanced version of NeRF that employs 3D conical frustums instead of single rays, resulting in anti-aliased and visually clear synthesized views. TensoRF utilizes tensor decomposition to parameterize the model as a collection of low-rank tensor components, thereby improving the quality of reconstruction. We report the qualitative and quantitative comparison to baseline algorithms in Table \ref{tab:results}. and Figure \ref{fig:compare}.. The experimental results indicate that our method achieves outstanding performance in generating high-quality images from novel viewpoints. These metrics are obtained by calculating the average value of multiple scenes in each dataset. We use the related open-source projects to implement these methods. Additionally, in Figure \ref{fig:motor}., we present a visual comparison between our method and the baseline on real-world datasets.

\begin{figure*}
    \centering
    \includegraphics[width=1\linewidth]{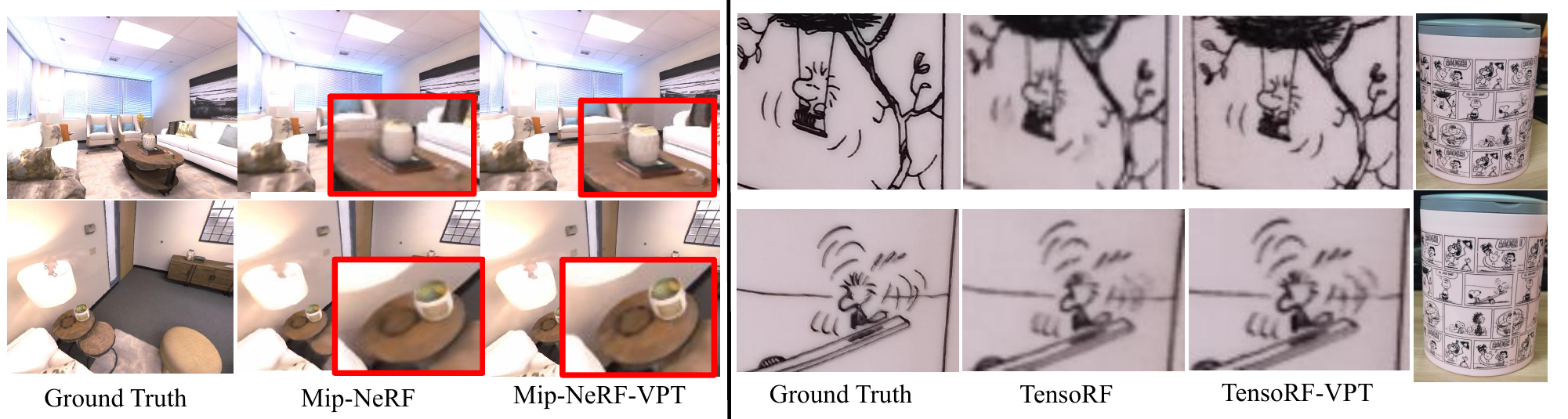}
    \caption{Qualitative evaluation of the improvement over Mip-NeRF and TensoRF on Replica dataset and user-captured dataset. Please zoom in for better viewing.}\label{fig:plug}
\end{figure*}

\begin{table}[tbp]
    \centering
    \bgroup
    \resizebox{1.0\linewidth}{!}{
    \small
        \begin{tabular}{c|cc|cc}
        \hline
        & \multicolumn{2}{c|}{Replica Dataset} & \multicolumn{2}{c}{User-Captured Dataset} \\
        & PSNR$\uparrow$ & SSIM$\uparrow$  & PSNR$\uparrow$ & SSIM$\uparrow$  \\
        \hline
            Mip-NeRF                  & 33.12 & 0.961 & 26.73 & 0.841 \\
            Mip-NeRF-VPT              & 35.63     & 0.983   & 28.55 & 0.855 \\
            \hline
            TensoRF                  & 33.17 &  0.969 & 27.01 & 0.838 \\
            TensoRF-VPT              & 34.62  & 0.977 & 28.89 & 0.853       \\
        \hline
        \end{tabular}
    }
    \egroup
    \caption{Quantitative comparison of the results obtained 
    by applying view prompt tuning on Mip-NeRF and TensoRF.
    }
    \label{tab:plug}
\end{table}

\subsection{Plug and Play} 
Our method is based on the framework of NeRF, essentially trying to improve the learning ability of MLPs for the scene via visual prompt tuning and reusing the model. For NeRF and a series of NeRF-based works, we can incorporate the view prompt of the model to improve the view synthesis quality through iterative methods. To fully verify the generalization ability of our \abbr, we conducted experiments not only with the NeRF baseline but also integrated our \abbr\ into other renowned methods such as Mip-NeRF and TensoRF. Based on Mip-NeRF and TensoRF, we add the view prompt to the network and iteratively learn the model. Our objective is to improve the quality of synthesized images based on these two methods. Figure \ref{fig:plug}. and Table \ref{tab:plug}. show our experimental results. We find our method raises NeRF-based model performance entirely to new levels via visual prompt tuning.

\subsection{Adaptability to Sparse-View Novel View Synthesis}

\begin{figure}
    \centering
    \footnotesize
    \includegraphics[width=\linewidth]{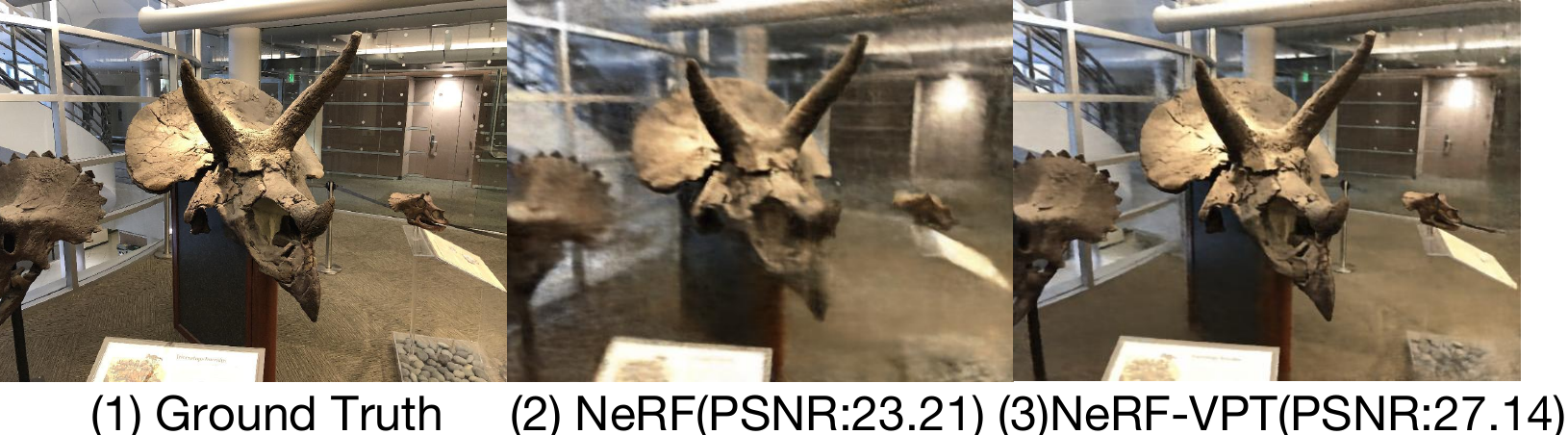}
    \caption{Visual comparisons in sparse view between NeRF and NeRF-VPT on forward-facing datasets.}\label{fig:horn}
\end{figure}

Neural Radiance Fields rely solely on RGB supervision information for comprehending the entire scene and generating images based on the viewing angle during the 3D reconstruction process. Consequently, it exhibits a strong reliance on densely sampled viewing angles. To address this limitation, we propose incorporating prior knowledge during the reconstruction stage to facilitate the network's understanding of the scene. This approach reduces the network's difficulty in comprehending the scene and consequently diminishes its dependence on densely sampled views. We conducted a comparative analysis of image quality between NeRF rendering using the original 180 views and our method utilizing multiple sparse views on the Replica dataset. The results, presented in Table \ref{tab:view}., demonstrate that while training NeRF on 180 views and our method on 30, 45, and 60 views, the image quality achieved after training with approximately 45 views is comparable to NeRF training with 180 views. These results indicate that our method is capable of achieving a NeRF-like rendering effect even with sparser viewing angles, effectively emulating the dense viewing angle rendering capability of NeRF. In essence, our method enables the achievement of NeRF-like high-quality rendering effects under sparse views. In Figure \ref{fig:horn}., we also present a visual comparison result of NeRF and NeRF-VPT under sparse views in the forward-facing dataset.

\subsection{Ablation Study}

\begin{figure}
    \centering
    \includegraphics[width=\linewidth]{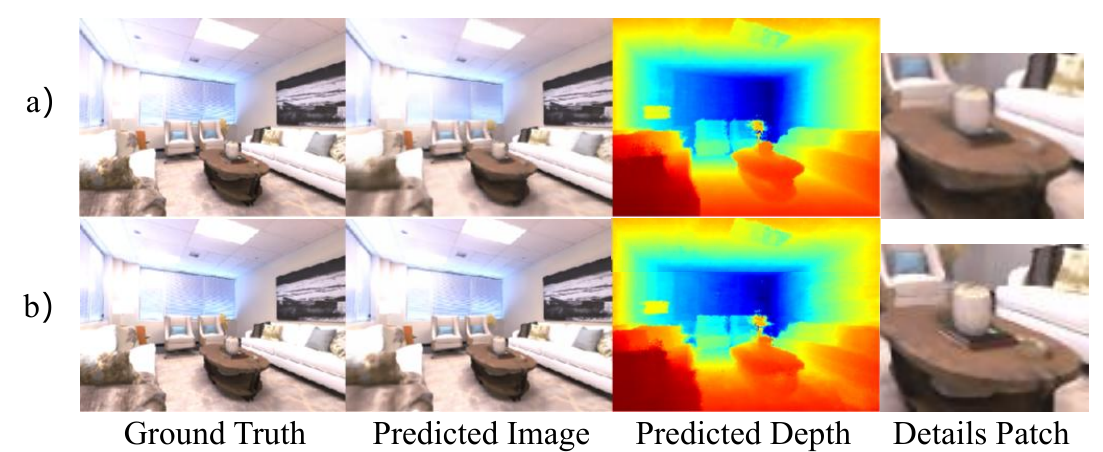}
    \caption{Qualitative comparison of different methods on the depth map. a) NeRF; b) NeRF-VPT. From left to right are the real image, the predicted image, the depth map, and the details of the predicted image. It can be seen from the figure that while improving the quality of the view synthesis, we also maintain the density of the reconstruction.}\label{fig:depth}
    \vspace{-10pt}
\end{figure}

\textbf{Depth Comparison.} While a notable enhancement in image quality, \abbr\ gains limited improvement in depth estimation. As illustrated in Figure \ref{fig:depth}. and Table \ref{tab:depth}., 
the improved image quality does not seem to result in any significant qualitative or quantitative improvements in predicted depth. We suspect the task of image synthesis differs significantly from that of depth prediction, with the former requiring a smooth distribution and the latter requiring a sharp distribution. Nevertheless, we believe that the cascade structure employed by \abbr\ has the potential to be extended to other methods in 3D reconstruction and therefore remains a promising avenue for future research. The Replica dataset provides a real depth map, so we make a quantitative analysis of the depth predicted by the reconstruction model on it. The calculation of the depth distance refers to the form of Mean-Squared Loss. We calculated the MSE of the predicted depth map and the predicted depth map according to the ray sampling.

\begin{table}
    \centering
    \bgroup
    \small
        \begin{tabular}{l@{\hskip 5.5pt}|@{\hskip 5.5pt}c@{\hskip 5.5pt}c@{\hskip 5.5pt}c@{\hskip 5.5pt}@{\hskip 5.5pt}c@{\hskip 3pt}c@{}}

        \hline
         & \multicolumn{2}{c@{\hskip 5.5pt}@{\hskip 5.5pt}}{Replica}  \\

         & PSNR$\uparrow$ & SSIM$\uparrow$  \\
        \hline
        \hline
            NeRF (180view)                  & 31.22 & 0.912    \\
            Ours Stage 1 (30view)                   & 30.31 & 0.896    \\
            Ours Stage 1 (45view)                   & 31.56 & 0.925   \\
            Ours Stage 1 (60view)                   & 32.87 & 0.950   \\
        \hline
        \end{tabular}
    \egroup
    \caption{
    Quantitative comparisons for the sampling views. We compared the reconstruction results of NeRF at 180 views and Ours at different views on the Replica dataset.
    }\label{tab:view}
\end{table}

\begin{table}
    \centering
    \bgroup
    \small
        \begin{tabular}{l@{\hskip 5.5pt}|@{\hskip 5.5pt}c@{\hskip 5.5pt}c@{\hskip 5.5pt}c@{\hskip 5.5pt}@{\hskip 5.5pt}c@{\hskip 3pt}c@{}}
        \hline
         & PSNR$\uparrow$ & depth distance\\
        \hline
        \hline
            NeRF               & 31.22 & 0.02    \\
            NeRF-VPT  & 34.95 & 0.02\\
        \hline
        \end{tabular}
    \egroup
    \caption{
    Quantitative comparisons for depth distance.
    We compared the depth performed by NeRF and NeRF-VPT on the Replica dataset, and the table shows that the predicted depth can be maintained while improving the image quality.
    }
    \label{tab:depth}
\end{table}

\begin{table}
    \centering
    \resizebox{0.7\linewidth}{!}{
    \begin{tabular}{c|c|c}
\hline
       &  Parameter Number     & FPS   \\
\hline
NeRF   &  1.1916 M  &  0.103    \\
Ours Stage 1  &  1.1924 M   &  0.108   \\
\hline
\end{tabular}
}
    \caption{Comparison on complexity with NeRF. FPS is the rendered frame per second.}\label{tab:complex_aly}
\end{table}

\begin{table}
    \centering
    \bgroup
    \small
    \resizebox{0.95\columnwidth}{!}{
        \begin{tabular}{l@{\hskip 5.5pt}|@{\hskip 5.5pt}c@{\hskip 5.5pt}|c@{\hskip 5.5pt}c@{\hskip 5.5pt}}
        \hline
         & View Prompt & PSNR$\uparrow$ & SSIM$\uparrow$ \\
        \hline
        \hline
        NeRF w/o prior & No Prior & 31.22  & 0.912  \\
        NeRF-VPT  & Feats & 31.76 & 0.919 \\
        NeRF-VPT dir. & Gaussian Noise & 11.58  & 0.567 \\
        NeRF-VPT dir. & Ground Truth & 55.83 & 0.977 \\
        NeRF-VPT pos. & Last Stage Images & 32.88 & 0.923 \\
        NeRF-VPT dir.& Last Stage Images & 33.96 & 0.936 \\
        \hline
        \end{tabular}
        }
    \egroup
    \caption{
    {Quantitative comparison of effect of prior knowledge.
     We show the comparative effects of adding prior knowledge to NeRF in different forms. Here, "Feats" means feature fusion as prior for training and testing; "dir. +"  means incorporating prior to the direction; "pos. +" means incorporating prior to the position. 
    }}
    \label{tab:piro_knowledge}
\end{table}

\begin{table}
    \centering
    \bgroup
    \small
        \begin{tabular}{l@{\hskip 5.5pt}|@{\hskip 5.5pt}l@{\hskip 5.5pt}l@{\hskip 5.5pt}c@{\hskip 5.5pt}@{\hskip 5.5pt}c@{\hskip 3pt}c@{}}

        \hline

         &  PSNR$\uparrow$ & SSIM$\uparrow$ & Time(h) \\
        \hline
         Stage 0 & 31.22 & 0.912 & 8.78 \\
         Stage 1 & 33.96 & 0.936 & 2.42 \\
	   Stage 2 & 34.76  & 0.943 & 1.96 \\ 
         Stage 3 & 34.99  & 0.947 & 1.54\\
         Stage 4 & 35.19  & 0.951 & 1.22 \\
         Stage 5 & 35.32  & 0.953 & 0.92\\
         Stage 6 & 35.39  & 0.954 & 0.61 \\
        \hline
        \end{tabular}
    \egroup
    \caption{The performance of our method at each stage and the corresponding time consumed. Conducting experiments on the replica dataset.}
    \label{tab:iteration}
\end{table}

\textbf{Computational Complexity.}
We compare the parameter number and the rendered frame per second (FPS) in Table \ref{tab:complex_aly}.. 
Our method brings negligible cost to networks, with only 0.0008 M overhead. As we adopt recurrent modules, it is inevitable that \abbr\ requires more time for training. Table \ref{tab:iteration}. outlines the training time of NeRF-VPT in each stage for 640 $\times$ 480 indoor scenes on Replica, showing our method is cost-effective: From Stage 1 to 6, the PSNR consistently increases, but the training time does not increase proportionally despite the visual prompts becoming more accurate and informative. As the iterations progress, the training time per round significantly decreases, primarily due to the increasing accuracy of the given visual prompts, which reduces the learning difficulty for the network.

\textbf{Effect of Prior Knowledge.} Prior knowledge plays a crucial role in our method, so how and where to incorporate the prior to warrant further study. Hence, we conducted a comparative analysis, which involved the following steps: i) Extracting and fusing features from scene images. We uniformly sampled 20 images from the training set to represent the entire scene and utilized the Swin-Transformer to extract their features. Subsequently, we employed soft-view pooling\cite{LeiLi2020EndtoEndLL} to fuse the extracted features. Finally, we integrated these fused features into each Linear Layer within the MLPs. ii) As discussed in the previous section on the method, we also explored the direct integration of ground truth as prior knowledge into the network. However, due to the unavailability of ground truth for unknown viewpoints during testing, we utilized the ground truth from the test set as prior in this set of experiments. Furthermore, we supplemented the results with experiments using Gaussian noise as a substitute for prior knowledge; iii)We also conducted a comparison to evaluate the disparities arising from incorporating prior knowledge in position encoding and direction encoding. The experimental results further validated our hypothesis: the ability of MLPs to predict color is more significantly influenced by the direction information. This can be attributed to the fact that, during the reconstruction of objects, after predicting volume density based on positional information, different direction information can induce diverse color variations for the same position. We did the above comparative experiments on the Replica dataset, and the detailed experimental results are in Table \ref{tab:piro_knowledge}.

\textbf{Effect of Prior Model Re-utilization.} 
Our method enhances the quality of synthesized images by incorporating prior knowledge as guidance during the iterative process. We illustrate this evolution in Table \ref{tab:iteration}, revealing that the performance reaches its peak after the sixth iteration stage.
\section{Conclusion}

This study proposes a novel and general framework to improve the performance of view synthesis based on NeRF. We present \abbr\, which introduces a new structure with recurrent modules and adopts the output of NeRF as priors. This allows \abbr\ to significantly improve the quality of view-dependent appearance. It is portable-friendly and is capable of combining with the existing methods to get state-of-art performance. We believe that this work provides new perspectives to take full advantage of representation.

\section{Acknowledgements}
This work was supported by the following grants: the National Key R \& D Program of China under Grant 2021ZD0111601, the National Natural Science Foundation of China (NSFC) under Grants 62276283, the Guangdong Basic and Applied Basic Research Foundation under Grants  2023A1515012985, the Basic and Applied Basic Research Special Projects under Grant SL2022A04J01685, and the UKRI grant of Turing AI Fellowship EP/W002981/1. We would also like to thank the Royal Academy of Engineering and FiveAI as well as the Guangdong Province Key Laboratory of Information Security Technology.
\bibliography{aaai24}

\clearpage
\maketitle

We provide additional materials that support our main experimental results.

\section{Algorithm}
We include an algorithm flowchart to aid in understanding the implementation of our method in Algorithm.~\ref{alg:nerf-vpt}.
In the first stage, we incorporate the images rendered by NeRF as view prompts into the network as inputs and proceed with the iterations. Throughout each round of training, the view prompts maintain the same perspective as the sampled views, providing prior knowledge to assist the network in understanding the scene. As the iterations progress, both the quality of the view prompts and the reconstructed results improve, achieving view prompt tuning. The view prompt for each iteration is derived from the output of the previous iteration, and the iterations are halted when the improvement in view prompt quality becomes negligible.

\begin{algorithm}[htbp]
\caption{NeRF-VPT training procedure}
\label{alg:nerf-vpt}
\begin{algorithmic}[1]
\STATE Input: position x, direction d
\STATE $(c, \sigma) = F(x, d)$
\STATE View Prompt: $C_0=\int_{t_n}^{t_f} Tc(t)\sigma(t) dx$
\STATE Stage 0: $(c, \sigma) = F(x, (d, C_0))$
\WHILE{$|C_i-C_{i-1}|>threshold$}
    \STATE Input: x, d, $C_i$
    \STATE View Prompt Tuning: $C_i=\int_{t_n}^{t_f}Tc(t)\sigma(t) dx$
    \STATE Stage i: $(c, \sigma) = F(x, (d, C_i))$
\ENDWHILE
\end{algorithmic}
\end{algorithm}

\section{User-captured Dataset}
\begin{figure*}
    \centering
    \includegraphics[width=1\linewidth]{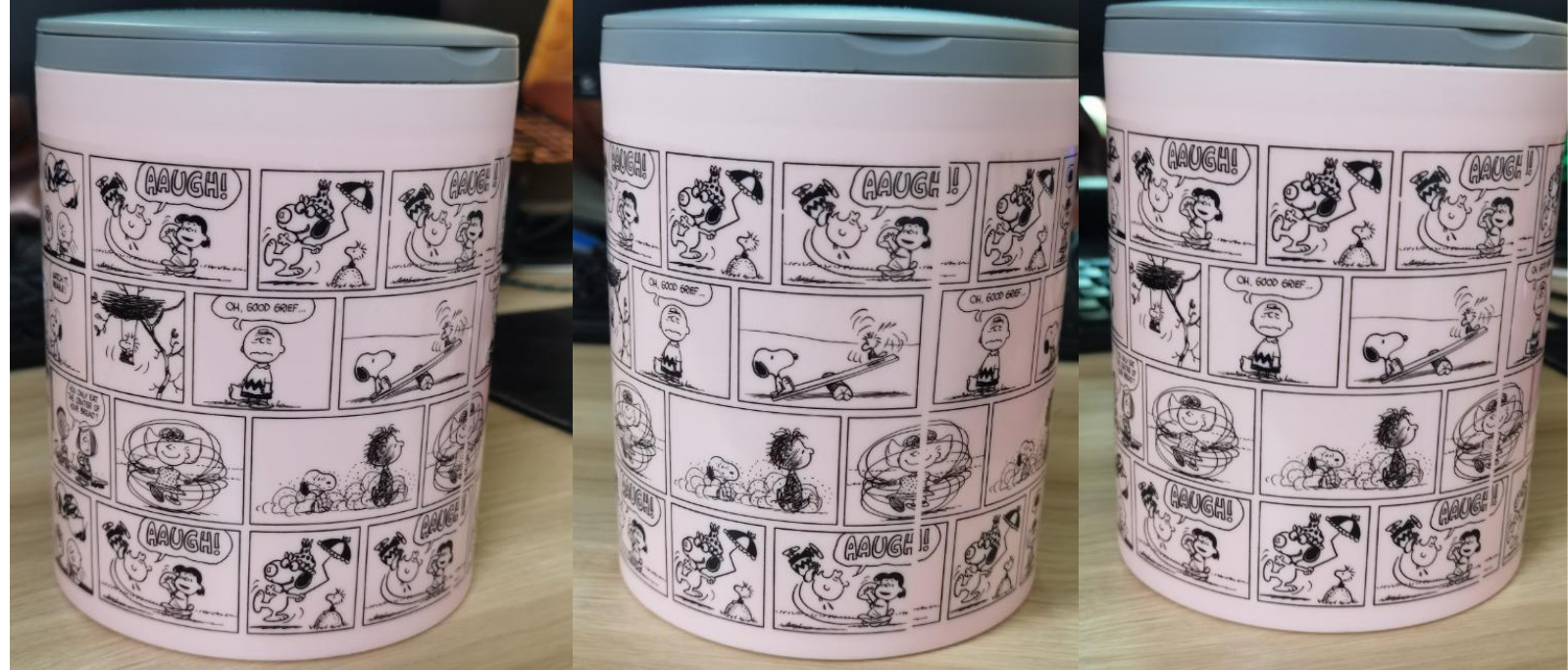}
    \caption{User-captured dataset with multi-angle images. Please zoom in for better viewing.}
    \label{fig:dataset}
\end{figure*}

In order to capture more intricate details during the reconstruction process, we collected a self-captured dataset. A total of 41 forward-facing images were captured with careful consideration given to various factors such as lighting conditions, camera settings, and scene coverage. The images were captured using a smartphone camera, and we utilized the COLMAP software to estimate the corresponding camera parameters for each image. The software effectively estimated the intrinsic and extrinsic camera parameters for each image in the dataset, including focal length, principal point, distortion coefficients, and camera orientations. The resulting dataset exhibits a diverse range of viewpoints, enabling comprehensive evaluation and analysis of our reconstruction approach. By utilizing self-captured images and accurate camera parameter estimation, we created a reliable dataset that aligns with the requirements and standards of similar datasets, such as the LLFF dataset~\cite{mildenhall2019local}, facilitating consistent and meaningful comparisons in subsequent reconstruction experiments. To maintain consistency with the experimental setup of the LLFF dataset, we resized the captured dataset to a resolution of 320x240 during the training process. Fig.~\ref{fig:dataset} illustrates our dataset. Please visit https://pan.baidu.com/s/1-V5RaYoUdeS6-ge-e291Sg?pwd=cq6g to view the dataset and extract the code is cq6g.

\begin{figure*}
    \centering
    \includegraphics[width=1\linewidth]{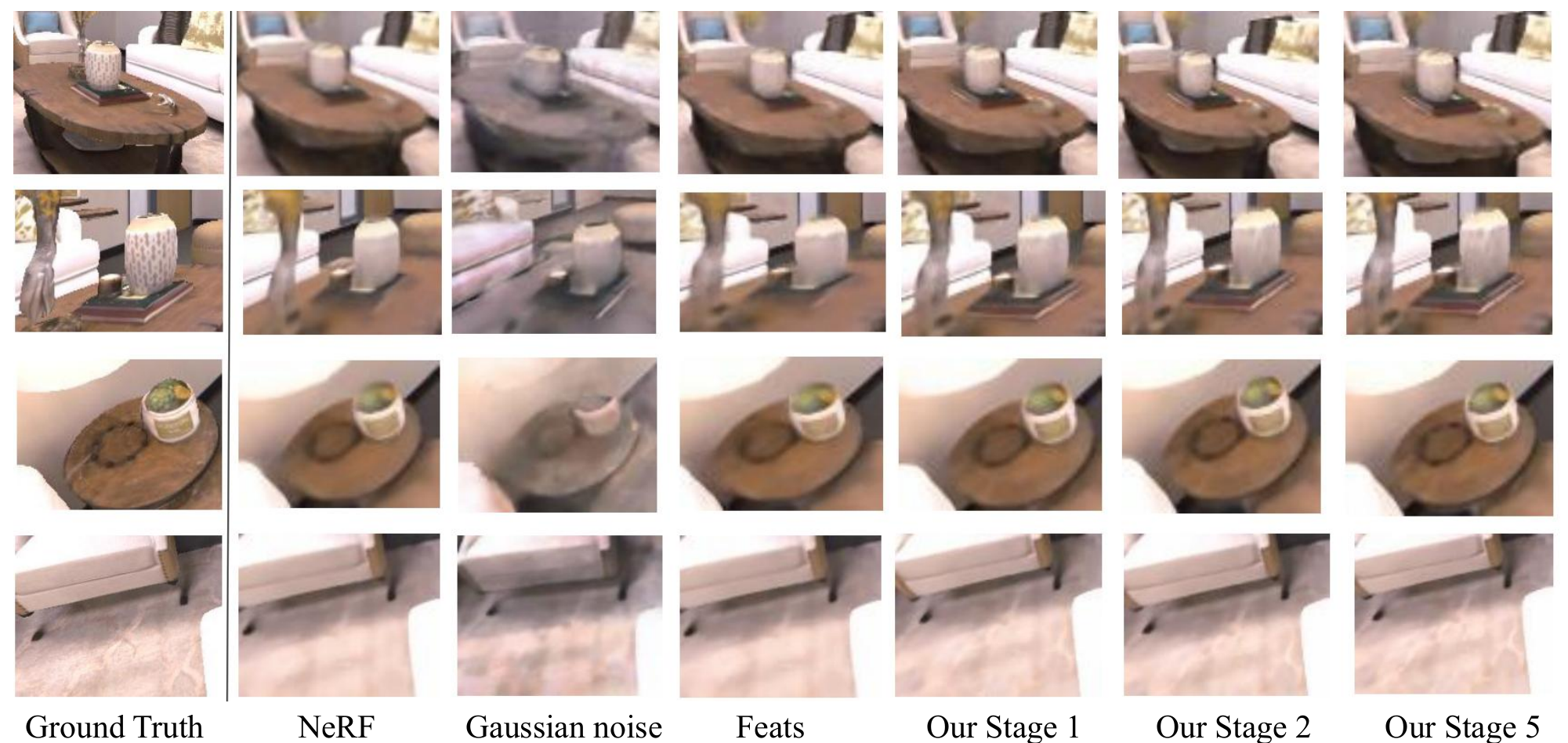}
    \caption{Visualization results about the effect of view prompt. 'Feats' represents the utilization of fused features as prior knowledge. 'Gaussian noise' represents the utilization of Gaussian noise as prior knowledge. Please zoom in for better viewing.}
    \label{fig:ablation}
\end{figure*}

\section{Additional Visualization Results}
\textbf{visualization results about the effect of view prompt}
We supplemented the visual information regarding the impact of the different view prompts mentioned. In the ablation experiments, we discussed the impact of incorporating different view prompts into our method, including 1) fused features, 2) Gaussian noise, 3) images rendered by the NeRF model, and 4) images rendered after view prompt tuning. We supplemented the visualization of these various prior knowledge effects in Fig.~\ref{fig:ablation}. The results indicate that the prior knowledge enhanced through view prompt tuning exhibited the greatest improvement in the quality of synthesizing novel view images.
\begin{figure*}
    \centering
    \includegraphics[width=1\linewidth]{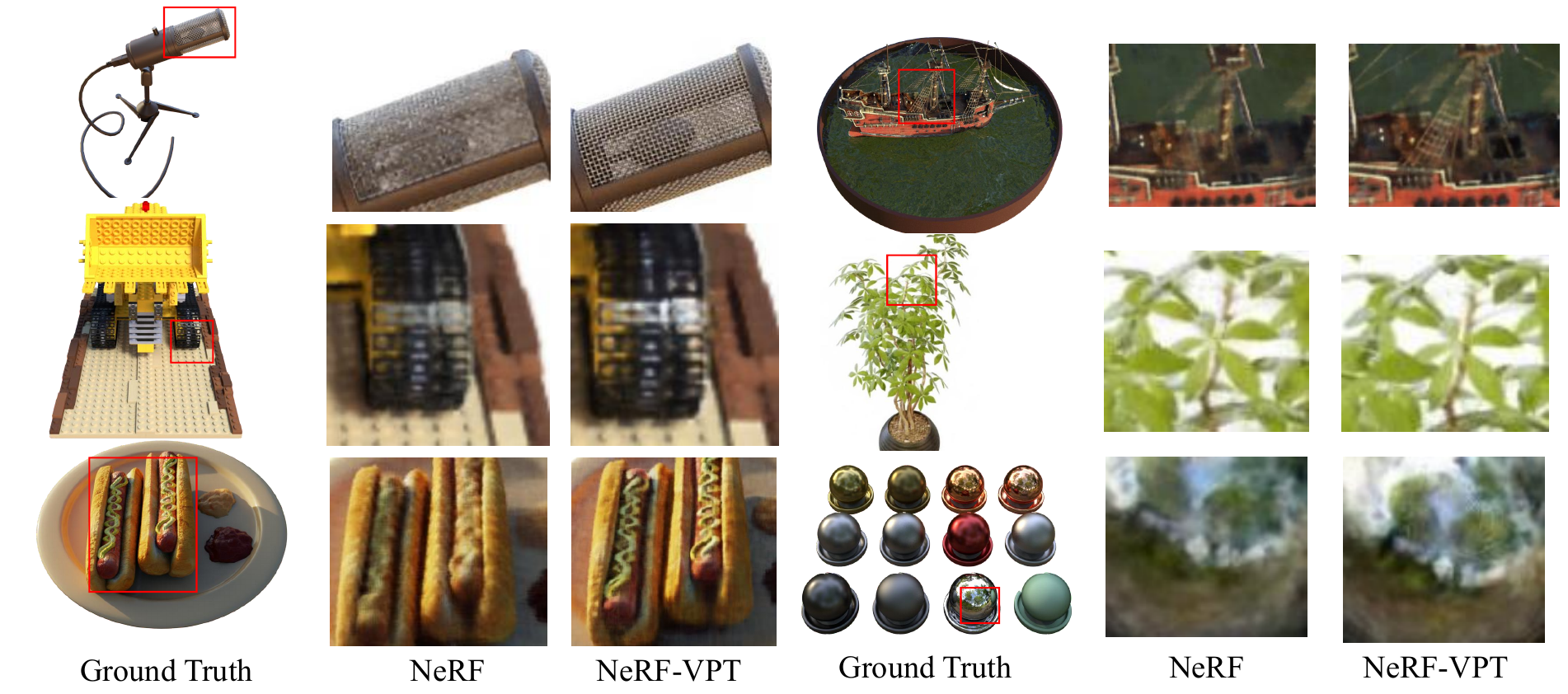}
    \caption{The visual results of our method compared to the baseline across multiple scenes.}
    \label{fig:visualization}
\end{figure*}

\begin{table*}
    \centering
    \vspace{-11pt}
    \resizebox{1.2\columnwidth}{!}{
       \begin{tabular}{l@{\hskip 3pt}|@{\hskip 3pt}l@{\hskip 3pt}l@{\hskip 3pt}c@{\hskip 3pt}@{\hskip 3pt}c@{\hskip 3pt}c@{\hskip 3pt}c@{\hskip 3pt}@{\hskip 3pt}c@{\hskip 3pt}c@{\hskip 3pt}c@{\hskip 3pt}@{\hskip 3pt}c@{\hskip 3pt}c@{\hskip 3pt}c@{}}
        \hline
         & chair & drums& ficus& hotdog& lego& materials& mic& ship \\
        \hline
        NeRF     & 33.00& 25.01& 30.13& 36.18& 32.54& 29.62& 32.91& 28.65 \\
        \hline
        NeRF-VPT           &  36.02 & 26.61  & 34.32 & 38.32 & 36.81  & 31.62  &  37.63 & 31.52   \\
        \hline
        \end{tabular}
    }
    \caption{
    {\bf Quantitative comparison.}
    We compare the PSNR metric between our method and the baseline in multiple scenarios to assess their performance differences.}
    \label{tab:psnr}
\end{table*}

\begin{table*}
    \centering
    \small
    \resizebox{1.2\columnwidth}{!}{
        \begin{tabular}{l@{\hskip 3pt}|@{\hskip 3pt}l@{\hskip 3pt}l@{\hskip 3pt}c@{\hskip 3pt}@{\hskip 3pt}c@{\hskip 3pt}c@{\hskip 3pt}c@{\hskip 3pt}@{\hskip 3pt}c@{\hskip 3pt}c@{\hskip 3pt}c@{\hskip 3pt}@{\hskip 3pt}c@{\hskip 3pt}c@{\hskip 3pt}c@{}}
        \hline
         & chair & drums& ficus& hotdog& lego& materials& mic& ship \\
        \hline
        NeRF     & 0.967 &0.925& 0.964& 0.974& 0.961& 0.949& 0.980& 0.856 \\
        \hline
        NeRF-VPT   &  0.989 & 0.941  & 0.991 & 0.993  & 0.987  & 0.971  &0.995   &  0.896  \\
        \hline
        \end{tabular}
    }
    \vspace{-8pt}
    \caption{
    {\bf Quantitative comparison.}
    We compare the SSIM metric between our method and the baseline in multiple scenarios to assess their performance differences.}
    \label{tab:ssim}
    \vspace{-1em}
\end{table*}

\textbf{visualization results about more detail}
We supplement additional visualization results in Fig.~\ref{fig:visualization}, showcasing the improved details of our method compared to the baseline across various scenes. Additionally, we report the corresponding PSNR and SSIM ~\cite{wang2004image}  metrics for our method and the baseline in Table.~\ref{tab:psnr} and Table.~\ref{tab:ssim}, respectively, for each scene.


\end{document}